\documentclass[9.5pt,cspaper,journal,compsoc]{IEEEtran}
\usepackage{multirow}
\usepackage{booktabs}
\usepackage{float}
\usepackage{color}
\ifCLASSOPTIONcompsoc
  \usepackage[nocompress]{cite}
\else
  \usepackage{cite}
\fi
\ifCLASSINFOpdf
  \usepackage[pdftex]{graphicx}
\else
  \usepackage[dvips]{graphicx}
\fi
\usepackage{amsmath}
\usepackage{amssymb}
\usepackage{url}

\begin{document}
%
\title{Disentangled Variational Autoencoder for Emotion Recognition in Conversations}
%
%
%
%

\author{Kailai~Yang,~\IEEEmembership{}
        Tianlin~Zhang,~\IEEEmembership{}
        Sophia~Ananiadou~\IEEEmembership{}
\IEEEcompsocitemizethanks{\IEEEcompsocthanksitem Kailai Yang, Tianlin Zhang, and Sophia Ananiadou are with NaCTeM and the
Department of Computer Science, The University of Manchester,  United
Kingdom.\protect\\
E-mail: kailai.yang@postgrad.manchester.ac.uk\\
E-mail: tianlin.zhang@postgrad.manchester.ac.uk\\
E-mail: sophia.ananiadou@manchester.ac.uk
}
}
%
%

\markboth{Journal of \LaTeX\ Class Files,~Vol.~14, No.~8, August~2015}%
{Shell \MakeLowercase{\textit{et al.}}: Bare Demo of IEEEtran.cls for Computer Society Journals}
%



\IEEEtitleabstractindextext{%
\begin{abstract}
In Emotion Recognition in Conversations (ERC), the emotions of target utterances are closely dependent on their context. Therefore, existing works train the model to generate the response of the target utterance, which aims to recognise emotions leveraging contextual information. However, adjacent response generation ignores long-range dependencies and provides limited affective information in many cases. In addition, most ERC models learn a unified distributed representation for each utterance, which lacks interpretability and robustness. To address these issues, we propose a \textbf{VAD}-disentangled \textbf{V}ariational \textbf{A}uto\textbf{E}ncoder (VAD-VAE), which first introduces a target utterance reconstruction task based on Variational Autoencoder, then disentangles three affect representations Valence-Arousal-Dominance (VAD) from the latent space. We also enhance the disentangled representations by introducing VAD supervision signals from a sentiment lexicon and minimising the mutual information between VAD distributions. Experiments show that VAD-VAE outperforms the state-of-the-art model on two datasets. Further analysis proves the effectiveness of each proposed module and the quality of disentangled VAD representations. The code is available at \url{https://github.com/SteveKGYang/VAD-VAE}.
\end{abstract}

}

\maketitle

\IEEEdisplaynontitleabstractindextext

%
\IEEEpeerreviewmaketitle

\IEEEraisesectionheading{\section{Introduction}\label{sec:introduction}}

%
%
%
%
\IEEEPARstart{E}{motion} Recognition in Conversations (ERC) aims to identify the emotion of each utterance within a dialogue from pre-defined emotion categories~\cite{8764449}. As an extension of traditional emotion detection from text, ERC attracts increasing research interest from the NLP community since it is more suitable for usage in real-world scenarios. For example, ERC aids dialogue systems in generating emotionally coherent and empathetic responses~\cite{MA202050}. It has also been widely utilised in emotion-related social media analysis~\cite{nandwani2021review, CHATTERJEE2019309} and opinion mining from customer reviews~\cite{9454192,Wang_Wang_Sun_Li_Liu_Si_Zhang_Zhou_2020}. 

Unlike sentence-level emotion recognition, the emotion of each utterance is dependent on contextual information in ERC. Some works enhance the context modelling ability by training the model to reconstruct the dialogue. For example, \cite{hazarika2021conversational,chapuis-etal-2020-hierarchical} pre-train the utterance encoders on large-scale conversation data and transfer the pre-trained weights to ERC. More recent works utilise Pre-trained Language Models (PLMs)~\cite{qiu2020pre} to model the dialogue and avoid pre-training from scratch~\cite{Shen_Chen_Quan_Xie_2021,xie-etal-2021-knowledge-interactive}. \cite{li2021contrast} combine both methods by fine-tuning pre-trained BART~\cite{lewis-etal-2020-bart} with an auxiliary response generation task on the dialogue, which trains the model to generate the next sentence given the target utterance. This task aims to force the model to recognise emotions considering context information.

\begin{figure}[htpb]
\centering
\includegraphics[width=8cm,height=5.391cm]{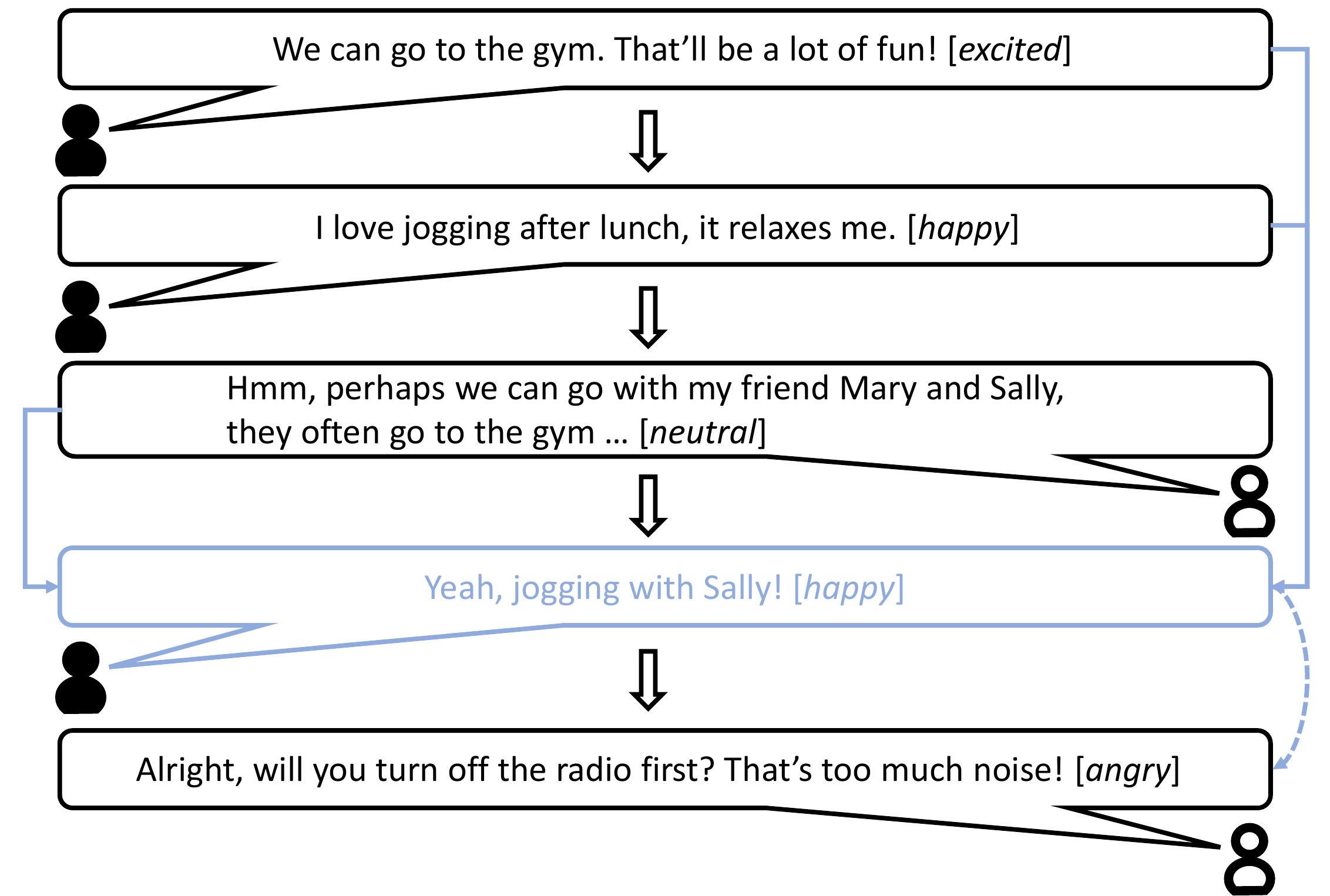}
\caption{A dialogue example. Solid lines show the influence of previous utterances on the emotion of the target utterance (marked blue), and dashed lines denote the change of the discussion topic in next utterance. As illustrated, the emotion of the target utterance is dependent on long-range history. As the topic changes, response generation provides limited affective information.}
\label{fig:example}
\end{figure}

However, response generation only mines the dependencies between two adjacent utterances, while the influence of long-range history on the target utterance is ignored. Generating the next sentence also provides limited affective information for the target utterance in many cases, such as the sudden change of the discussion topic. An example is shown in Figure \ref{fig:example} to prove the above points. As illustrated, the emotion of the target utterance is dependent on long-range history, and response generation provides limited affective information as the discussion topic changes in the next utterance. We argue that a context-aware reconstruction of the target utterance itself is more appropriate since ERC is centred on the target utterance representations. In addition, current ERC methods mostly learn a unified distributed representation for each target utterance. Though achieving impressive results, entangled features are proved to lack interpretability and robustness~\cite{6472238}. Affective text generation models~\cite{MA202050, goswamy-etal-2020-adapting} also postulate that emotions are independent of the content they modify, and their success shows the viability of disentangling emotion features from content representations. 


To address these issues, we propose a \textbf{VAD}-disentangled \textbf{V}ariational \textbf{A}uto\textbf{E}ncoder (VAD-VAE) for ERC. Firstly, instead of generating the response, we introduce a target utterance reconstruction task based on the Variational Autoencoder (VAE)~\cite{DBLP:journals/corr/KingmaW13} generative model. We devise a PLM-based context-aware encoder to model the dialogue, and sample the latent representations from a Gaussian distribution estimated from the utterance representations. The Gaussian distribution also aims to regularise the latent space. Then another PLM-based decoder is leveraged to reconstruct the target utterance from the latent representations. VAD-VAE outperforms the state-of-the-art model on two ERC datasets.

Secondly, we utilise disentangled representation learning~\cite{higgins2018towards} techniques to disentangle critical features from the utterance representations. Studies in affect representation models in psychology point out that Valence-Arousal-Dominance (VAD) are both orthogonal and bipolar, which are appropriate to define emotion states~\cite{RUSSELL1977273,mehrabian1995framework}. Therefore, we propose to disentangle the three VAD features from the latent space of the VAE, where we also sample each of the VAD representations from a corresponding Gaussian distribution estimated from the utterance representations. Then the disentangled features are combined for both ERC and target utterance reconstruction tasks.

Thirdly, two techniques are used to enhance the disentangled VAD representations. We boost their \emph{informativeness}~\cite{eastwood2018framework} by introducing supervision signals from NRC-VAD~\cite{mohammad-2018-obtaining}, a sentiment lexicon that contains human ratings of VAD for all emotions. To enforce the \emph{independence}~\cite{higgins2018towards} of latent spaces, we minimise the Mutual Information (MI) between VAD representations. During training, we estimate and minimise the variational Contrastive Log-ratio Upper-Bound (vCLUB)~\cite{pmlr-v119-cheng20b} of MI. Further analysis proves the quality of the disentangled VAD representations.

To summarise, this work mainly makes the following contributions:
\begin{itemize}
\item We propose a VAE-based target utterance reconstruction auxiliary task for ERC, which improves model performance and regularises the latent spaces.
\item For the first time in ERC, We explicitly disentangle the three VAD features from the utterance representations. Analysis shows it benefits interpretability and robustness, and bears potential in the affective text generation task.
\item We enhance the \emph{informativeness} of the disentangled representations with VAD supervision signals from the lexicon NRC-VAD and minimise the vCLUB estimate of their mutual information to improve \emph{independence}.
\end{itemize}

\section{Related Work}
\subsection{Emotion Recognition in Conversations} 
For ERC, the emotion of a dialogue participant is primarily influenced by the dialogue history, which makes context modelling a key challenge. Early works utilised Recurrent Neural Networks (RNN) to model each participant's dialogue flow as a sequence and revise them as memories at each time step~\cite{hazarika-etal-2018-icon,hazarika-etal-2018-conversational}. Considering multi-party relations, \cite{majumder2019dialoguernn} leveraged another global-state RNN to model inter-speaker dependencies and emotion dynamics. To avoid designing complex model structures, more recent works leveraged the strong context-modelling ability of PLMs to model the conversation as a whole~\cite{li-etal-2020-hitrans,Shen_Chen_Quan_Xie_2021}. Some other works~\cite{shen-etal-2021-directed,li-etal-2021-past-present} built a graph upon the dialogue with each utterance as a node and used graph neural networks to model ERC as a node-classification task.

Enhancing the utterance representations is also crucial for ERC. Some works managed to incorporate task-related information. For example, commonsense knowledge was introduced~\cite{ghosal-etal-2020-cosmic,xie-etal-2021-knowledge-interactive} to enrich the semantic space. To enhance the conversation modelling ability, some methods pre-trained the model on large-scale conversation data and transferred the weights to ERC~\cite{chapuis-etal-2020-hierarchical,hazarika2021conversational}. Multi-task learning was also leveraged to introduce topic information~\cite{wang2020sentiment,zhu-etal-2021-topic}, discourse roles~\cite{DBLP:conf/aaai/OngSCLNLSLW22} and speaker-utterance relations~\cite{li-etal-2020-hitrans} to aid emotion reasoning. \cite{park-etal-2021-dimensional,10.1145/3404835.3463080} incorporated VAD information to introduce fine-grained sentiment supervision. Contrastive learning~\cite{li2021contrast,10040720,song-etal-2022-supervised} was also devised to distinguish utterances with similar emotions.

\subsection{Disentangled Representation Learning} 
Disentangled Representation Learning (DRL) aims to map key data features into distinct and independent low-dimensional latent spaces~\cite{higgins2018towards}. Current DRL methods are mainly divided into unsupervised and supervised disentanglement. Early unsupervised methods mainly designed constraints on the latent space to enforce the independence of each dimension, such as information capacity~\cite{burgess2018understanding} and mutual information gap~\cite{chen2018isolating}. Supervised methods introduced supervision signals to different parts of the latent space to enforce \emph{informativeness}. Some works utilised ground-truth labels of the corresponding generative factors, such as syntactic parsing trees~\cite{bao-etal-2019-generating} and style labels~\cite{cheng-etal-2020-improving}. In contrast, other works used weakly-supervised signals, including pairwise similarity between representations ~\cite{Chen_Batmanghelich_2020} and semi-supervised ground-truth labels~\cite{vasilakes-etal-2022-learning}. Still, supervised methods devised techniques such as mutual information minimisation~\cite{vasilakes-etal-2022-learning} and adversarial learning~\cite{bao-etal-2019-generating,john-etal-2019-disentangled} to enforce \emph{independence} and \emph{invariance}~\cite{DBLP:conf/iclr/ShuCKEP20} of the disentangled representations.

\section{Methodology}
\subsection{Task Definition}
The ERC task is defined as follows: a dialogue $D$ contains $n$ utterances $\{u_1, u_2,...,u_n\}$, with the corresponding ground-truth emotion labels $\{y_1, y_2,...,y_n\}$, where $y_i\in E$, $E$ is the pre-defined emotion label set. Each $u_i$ contains $m_i$ tokens: $\{u_i^1, u_i^2,...,u_i^{m_i}\}$. The dialogue is also accompanied by a speakers list $S(D)=\{S(u_1), S(u_2),...,S(u_n)\}$, where $u_i$ is uttered by $S(u_i)\in S$, and $S$ is the set of dialogue participants. With the above information, ERC aims to identify the emotion of each target utterance $u_i$, which is formalised as: $\hat{y}_i=f(u_i, D, S(D))$.

\subsection{Target Utterance Reconstruction}
\begin{figure}[htpb]
\centering
\includegraphics[width=7cm,height=8.8cm]{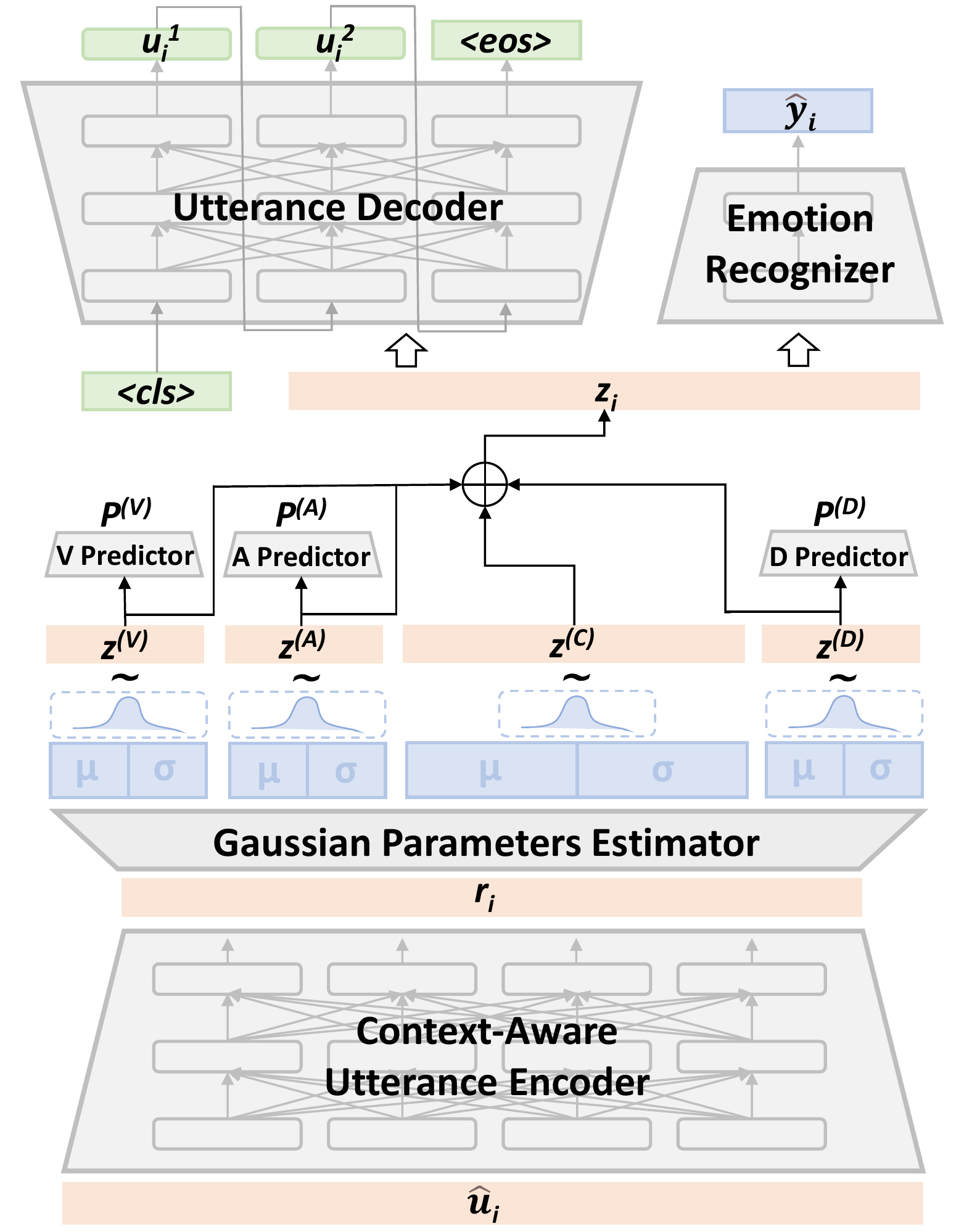}
\caption{Main components of VAD-VAE. Each latent representation $z^{(\mathcal{R})}$ is sampled from a Gaussian distribution estimated from the context-aware utterance representation $r$. The VAD prediction $P^{(\mathcal{R})}$ is obtained from $z^{(\mathcal{R})}$ via the VAD predictors. The concatenated representation $z$ is utilised for both ERC and target utterance reconstruction. In the utterance decoder, ``$\langle cls \rangle$'' denotes the start-of-sentence token, and ``$\langle eos \rangle$'' denotes the end-of-sentence token.}
\label{fig:vae}
\end{figure}

This section introduces the target utterance reconstruction auxiliary task. Based on the context-aware utterance encoder, we also disentangle VAD latent representations from the utterance representations and build a VAE-based generative model to reconstruct the target utterance, which is the backbone of VAD-VAE, as illustrated in Figure \ref{fig:vae}. 

\subsubsection{Context-Aware Utterance Encoder}
To introduce speaker information explicitly, we prepend the speaker $S(u_j)$ to each utterance: $u_j=\{S(u_j),u_j^1,u_j^2,...,u_j^k\}$, where $u_j^k$ is the $k$-th token of $u_j$. Then the target utterance $u_i$ is concatenated with both past and future dialogues to obtain the context-aware input $\hat{u}_i$:
$$\hat{u}_i = \{\langle cls\rangle;u_{i-W_p};...;\langle sep\rangle;u_i;\langle sep\rangle;...;u_{i+W_f};\langle eos\rangle\}$$
where $\{;\}$ denotes concatenation, $W_p$ and $W_f$ denotes the past and future context window size, $\langle cls\rangle$ and $\langle eos\rangle$ denote the start-of-sentence and end-of-sentence token. Two $\langle sep\rangle$ tokens are added at the start and end place of $u_i$ to identify the target utterance. Utilising a PLM-based encoder, we obtain the context-aware utterance embeddings:
\begin{equation}
    r_i = Encoder(\hat{u}_i)
\end{equation}
where $Encoder$ denotes the RoBERTa-Large~\cite{liu2019roberta} utterance encoder, $r_i\in \mathbb{R}^{L\times D_h}$ is the utterance representations, $L$ denotes the sequence length, and $D_h$ is the hidden states dimension. We leverage the embedding of the start-of-sentence token at position 0: $r_i^{[CLS]}\in \mathbb{R}^{D_h}$ as the utterance-level representation of $u_i$.

\subsubsection{VAE-based Generative Model}
We build a VAE-based generative model and disentangle three latent features, Valence-Arousal-Dominance (VAD) from the utterance representation, where Valence reflects the pleasantness of a stimulus, Arousal reflects the intensity of emotion provoked by a stimulus, and Dominance reflects the degree of control exerted by a stimulus~\cite{Warriner2013NormsOV}. We also define a “Content'' feature that controls the content generation of the target utterance.

A VAE is utilised to estimate this model, which imposes a standard Gaussian prior distribution on each latent space $Z$. The deterministic utterance representation is replaced with an approximation
of the posterior $q_{\phi}(z|x)$, which is parameterised by a neural network. We utilise four feed-forward neural networks to map $x=r_i^{[CLS]}$ to four sets of Gaussian distribution parameters $(\mu,\sigma)$, which parameterise the latent distributions of Valence, Arousal, Dominance, and Content, denoted as $\mathcal{R}\in\{V, A, D, C\}$. For each feature, we sample the latent representation $z^{(\mathcal{R})}$ from the Gaussian distribution defined by the corresponding $(\mu^{(\mathcal{R})},\sigma^{(\mathcal{R})})$ using the re-parameterisation trick~\cite{DBLP:journals/corr/KingmaW13}:
\begin{equation}
    z^{(\mathcal{R})}_i = \mu^{(\mathcal{R})}\odot \sigma^{(\mathcal{R})} + \epsilon \sim \mathcal{N}(\mathbf{0}, \mathbf{I})
\end{equation}
where $z^{(\mathcal{R})}_i\in \mathbb{R}^{D_{(\mathcal{R})}}$, $D_{(\mathcal{R})}$ is the pre-defined latent space dimension. Then the latent representations are concatenated: $z_i=[z^{V}_i; z^{A}_i; z^{D}_i; z^{C}_i]$. $z_i$ is used to initialise the decoder and reconstruct the target utterance:
\begin{equation}
    u_i^j = Softmax(Decoder(z_i, u_i^{<j}))
\end{equation}
where $Softmax$ denotes the softmax operation, $u_i^j$ denotes the $j$-th generated tokens, and $u_i^{<j}$ denotes the previously generated tokens. We utilise BART-Large decoder~\cite{lewis-etal-2020-bart} as $Decoder$, since it shares the vocabulary with RoBERTa-Large in Huggingface~\cite{wolf2019huggingface} implementations and is proved powerful in many generative tasks. As in standard VAE, we include a KL-divergence term for each latent space to keep the approximate posterior close to the prior distribution. During training, we utilise the Evidence Lower BOund (ELBO) as the training objective:
\begin{equation}
\begin{split}
    \mathcal{L}_{ELBO}(\phi, \theta)=-\mathbb{E}_{q_{\phi}(z_i|x)}\left[ log\  p_\theta(x|z_i)\right]+\\
    \sum_{\mathcal{R}\in\{V,A,D,C\}}\alpha_{\mathcal{R}}KL\left[q_{\phi}^{(\mathcal{R})}(z_i^{(\mathcal{R})}|x)||p(z_i^{(\mathcal{R})})\right]
\end{split}
\end{equation}
where $\phi$ and $\theta$ denote the parameters of the encoder and decoder, each $\alpha_{\mathcal{R}}$ weights the corresponding KL-divergence term, and standard Gaussian prior is used for each $p(z_i^{(\mathcal{R})})$.

\subsection{Enhancing VAD Representations}
We aim to enhance the disentangled VAD representations considering the following two aspects: (a). \emph{informativeness}: the representation should include enough information to predict the corresponding generative factor well~\cite{Higgins2017betaVAELB,eastwood2018framework}. (b). \emph{Independence}: for each generative factor, the representation should lie in an independent latent space~\cite{higgins2018towards}. Therefore, we introduce supervision signals from a sentiment lexicon to enforce \emph{informativeness} and a mutual information minimisation objective to enforce \emph{independence}.

\subsubsection{Informativeness}
To enhance the representation's ability to predict the corresponding generative factor, we introduce supervision signals from NRC-VAD~\cite{mohammad-2018-obtaining}, a VAD sentiment lexicon that contains reliable human ratings of VAD for 20,000 English terms. All NRC-VAD terms denote or connote emotions and are selected from commonly used sentiment lexicons and tweets. Each term is strictly annotated via best-worst scaling with crowdsourcing annotators, and an aggregation process calculates the VAD for each term ranging from 0 to 1. For example, the emotion \emph{happiness} is assigned $vad_{happiness} = \{0.960, 0.732, 0.850\}$. More details about NRC-VAD are in Sec. \ref{datasets}.

With the pre-defined categorical emotion set $E$, we extract the VAD score $vad_{e_j}=\{vad_{e_j}^V, vad_{e_j}^A, vad_{e_j}^D\}$for each of the emotion $e_j\in E$ from NRC-VAD, where $j\in [1, |E|]$. Since fine-grained VAD supervision signals are introduced, we expect to improve both the \emph{informativeness} of VAD representations and model performance on ERC.

Specifically, for each $\hat{\mathcal{R}}\in \{V, A, D\}$, we compute the corresponding prediction from the latent representation using a feed-forward neural network predictor, and a sigmoid function is utilised to map the prediction to the range $(0, 1)$:
\begin{equation}
    P_i^{(\hat{\mathcal{R}})}=\frac{1}{1+\emph{e}^{-(z_i^{(\mathcal{\hat{R}})}W^{(\mathcal{\hat{R}})}+b^{(\mathcal{\hat{R}})})}}
\end{equation}
where $W^{(\hat{\mathcal{R}})}$ and $b^{(\hat{\mathcal{R})}}$ are parameters of the predictor corresponding to $\hat{\mathcal{R}}$. As the training objective, we compute the mean squared error loss between the predictions and the supervision signals:
\begin{equation}
    \mathcal{L}_{INFO}(\phi, \lambda)= \frac{1}{N}\sum_{i=1}^N\sum_{\mathcal{\hat{R}}\in\{V,A,D\}}(P_i^{(\mathcal{\hat{R}})}-vad_{y_i}^{(\mathcal{\hat{R}})})^2
\end{equation}
where $\phi$ and $\lambda$ denote the parameters of the encoder and the predictor, $y_i$ denotes the emotion label of $i$-th utterance, and $N$ denotes the batch size.

\subsubsection{Independence}
We improve the independence of all disentangled latent spaces by making their distributions as dissimilar as possible. A common method is to minimise the Mutual Information (MI)~\cite{DBLP:conf/icml/PooleOOAT19} between each pair of latent variables: $\mathbf{I}(\mathcal{\hat{R}}_i;\mathcal{\hat{R}}_j)$, which is defined as: 
\begin{equation}
\begin{split}
    \mathbf{I}(\mathcal{\hat{R}}_i;\mathcal{\hat{R}}_j) = \mathbb{E}_{p(\mathcal{\hat{R}}_i,\mathcal{\hat{R}}_j)}\Big[log\frac{p(\mathcal{\hat{R}}_i, \mathcal{\hat{R}}_j)}{p(\mathcal{\hat{R}}_i)p(\mathcal{\hat{R}}_j)}\Big]
\end{split}
\end{equation}
where $\mathcal{\hat{R}}_i$, $\mathcal{\hat{R}}_j\in \{V, A, D\}$ and $i\neq j$. However, MI is hard to calculate in high-dimensional spaces. The conditional distribution between each pair of latent variables is also unavailable in our cases. Therefore, we utilise the variational Contrastive Log-ratio Upper-Bound (vCLUB) proposed by Cheng et al.~\cite{pmlr-v119-cheng20b} to estimate and minimise an upper bound of MI, which is calculated as follows:
\begin{equation}
\begin{split}
    \mathbf{I}_{vCLUB}(\mathcal{\hat{R}}_i;\mathcal{\hat{R}}_j) := \mathbb{E}_{p(\mathcal{\hat{R}}_i,\mathcal{\hat{R}}_j)}[log\ q_\delta(\mathcal{\hat{R}}_j|\mathcal{\hat{R}}_i)]-\\
    \mathbb{E}_{p(\mathcal{\hat{R}}_i)}\mathbb{E}_{p(\mathcal{\hat{R}}_j)}[log\ q_\delta(\mathcal{\hat{R}}_j|\mathcal{\hat{R}}_i)]
\end{split}\label{vclub1}
\end{equation}

A variational distribution $q_\delta(y|x)$ with parameter $\delta$ is used to approximate $p(y|x)$. In practice, we separately use a feed-forward neural network as an estimator to approximate the conditional distribution between each pair in VAD variables: $P(\mathcal{\hat{R}}_i|\mathcal{\hat{R}}_j)$ where $i\neq j$, and the parameters are updated along with VAD-VAE at each time step. Unbiased vCLUB estimates between each pair are summed to get the MI minimisation loss as the training objective:
\begin{equation}\label{vclub2}
\begin{split}
    \mathcal{L}_{MI}(\phi,\delta) = \frac{1}{N}\sum_{k=1}^N\sum_{i,j}\ \Big[log\ q_{\delta_{ij}}(z_k^{(\mathcal{\hat{R}}_i)}|z_k^{(\mathcal{\hat{R}}_j)})-\\
    \frac{1}{N}\sum_{l=1}^Nlog\ q_{\delta_{ij}}(z_l^{(\mathcal{\hat{R}}_i)}|z_k^{(\mathcal{\hat{R}}_j)})\Big]
\end{split}
\end{equation}
where $\delta_{ij}$ denotes the parameters of the corresponding estimator. The detailed proof of Eqn. \ref{vclub1} and \ref{vclub2} is available in \cite{pmlr-v119-cheng20b}. Since no extra supervision signals are introduced, we expect the model to still achieve comparable performance as a trade-off for more \emph{independence} of each latent space.




\subsection{Model Training}
For ERC task, the concatenated latent representation $z_i$ is utilised to compute the final classification probability:
\begin{equation}
    \hat{y}_i = Softmax(z_iW_0+b_0)
\end{equation}
where $W_0$ and $b_0$ are learnable parameters. Then we compute the ERC loss $\mathcal{L}_{ERC}$ using standard cross-entropy loss:
\begin{equation}
    \mathcal{L}_{ERC}=-\frac{1}{N}\sum_{i=1}^N\sum_{j=1}^{|E|}y_i^jlog\ \hat{y}_i^j
\end{equation}
where $y_i^j$ and $\hat{y}_i^j$ are $j$-th element of $y_i$ and $\hat{y}_i$. Finally, we combine all proposed modules and train in a multi-task learning manner:
\begin{equation}
    \mathcal{L}=\mathcal{L}_{ERC}+\mu_{E}\mathcal{L}_{ELBO}+\mu_{I}\mathcal{L}_{INFO}+\mu_{MI}\mathcal{L}_{MI}
\end{equation}
where the $\mu$s are the pre-defined weight coefficients.

\section{Experimental Settings}
\subsection{Datasets}\label{datasets}
We evaluate our model on the following three benchmark datasets, where the statistics are listed in Table \ref{tab:statistics}:

\begin{table}[h]
\caption{Statistics of the datasets. Conv. and Utter. denotes the conversation and utterance number. Utter./Conv denotes the average utterance number per dialogue.}
\label{tab:statistics}
\begin{center}
\setlength{\tabcolsep}{0.5mm}{
\begin{tabular}{lccc}
\toprule \textbf{Dataset} &  \textbf{Conv.(Train/Val/Test)} & \textbf{Utter.(Train/Val/Test)} &  \textbf{Utter./Conv} \\ \midrule
IEMOCAP & 100/20/31 &  4,778/980/1,622 & 49.2 \\
MELD &  1,038/114/280 & 9,989/1,109/2,610 & 9.6\\
DailyDialog & 11,118/1,000/1,000 & 87,170/8,069/7,740 & 7.9 \\
\bottomrule
\end{tabular}}
\end{center}
\end{table}

\textbf{IEMOCAP}~\cite{busso2008iemocap}: A two-party multi-modal conversation dataset derived from the scenarios in the scripts of the two actors. For all datatsets, we only utilise the text modality in our experiments. The pre-defined categorical emotions are \emph{neutral, sad, anger, happy, frustrated, excited}.

\textbf{MELD}~\cite{poria-etal-2019-meld}: A multi-party multi-modal dataset enriched from \emph{EmotionLines} dataset, collected from the scripts of American TV show \emph{Friends}. The pre-defined emotions are \emph{neutral, sad, anger, disgust, fear, happy, surprise}.

\textbf{DailyDialog}~\cite{li-etal-2017-dailydialog}: A dataset compiled from human-written daily conversations with only two parties involved and no speaker information. The pre-defined emotion labels are the Ekman’s emotion types: \emph{neutral, happy, surprise, sad, anger, disgust, fear}.

\begin{table}[h]
\caption{The NRC-VAD assignments to all emotions in the four datasets.}
\label{vads}
\begin{center}
\resizebox{.48\textwidth}{!}{
\begin{tabular}{l|ccccccc}
\toprule \textbf{IEMOCAP} & neutral & frustrated & sad & anger & excited & happy & --\\ \midrule
Valence & 0.469 & 0.060 & 0.052 & 0.167 & 0.908 & 0.960 & --\\
Arousal & 0.184 & 0.730 & 0.288 & 0.865 & 0.931 & 0.732 & --\\
Dominance & 0.357 & 0.280 & 0.164 & 0.657 & 0.709 & 0.850 & --\\
\midrule\midrule
\textbf{MELD} & neutral & joy & surprise & anger & sad & disgust & fear\\ \midrule
Valence & 0.469 & 0.980 & 0.875 & 0.167 & 0.052 & 0.052 & 0.073\\
Arousal & 0.184 & 0.824 & 0.875 & 0.865 & 0.288 & 0.775 & 0.840\\
Dominance & 0.357 & 0.794 & 0.562 & 0.657 & 0.164 & 0.317 & 0.293\\
\midrule\midrule
\textbf{DailyDialog} & neutral & anger & disgust & fear & happy & sad & surprise\\ \midrule
Valence & 0.469 & 0.167 & 0.052 & 0.073 & 0.960 & 0.052 & 0.875\\
Arousal & 0.184 & 0.865 & 0.775 & 0.840 & 0.732 & 0.288 & 0.875\\
Dominance & 0.357 & 0.657 & 0.317 & 0.293 & 0.850 & 0.164 & 0.562\\
\bottomrule
\end{tabular}}
\end{center}
\end{table}

We utilise information from \textbf{NRC-VAD}~\cite{mohammad-2018-obtaining} in our methodology, a sentiment lexicon that contains human ratings of Valence, Arousal, and Dominance for more than 20,000 English words. All the terms in NRC-VAD denote or connote emotions. Specifically, the terms are collected from sentiment lexicons such as NRC emotion lexicon~\cite{DBLP:journals/ci/MohammadT13}, General Inquirer~\cite{DBLP:conf/afips/StoneH63}, and ANEW~\cite{Warriner2013NormsOV}. These terms are first very strictly annotated via best-worst scaling with crowdsourcing annotators. Then an aggregation process calculates the VAD for each term ranging from 0 to 1. Take Valence as an example. The annotators are presented with four words at a time (4-tuples) and asked to select the word with the highest and lowest Valence. The questionnaire uses paradigm words that signify the two ends of the valence dimension. The final VAD scores are calculated from the responses: For each item, the score is the proportion of times the item was chosen as the best (highest V/A/D) minus the proportion of times the item was selected as the worst (lowest V/A/D). The scores were linearly transformed from interval 0 (lowest V/A/D) to 1 (the highest V/A/D). In NRC-VAD, the emotion prototypes of the labels for the three ERC datasets are listed in Table \ref{vads}. According to the assignments, most cluster centres reflect appropriate positions of the corresponding emotions in VAD space, where similar emotions are measurably closer to each other while maintaining a fine-grained difference to facilitate the model to distinguish them. For example, \emph{happy} stays closer to \emph{excited} than \emph{anger} in IEMOCAP. In addition, for all four datasets, positive and negative emotions are separated mainly by \emph{neutral} in the dimension Valence, while the emotions within each sentiment polarity mainly differ in Arousal and Dominance.

We also notice that human-labeled utterance-level VAD scores are available in IEMOCAP. Therefore, we further explore utilising these utterance-level VAD scores instead of NRC-VAD to supervise informativeness. The aggregation process calculates the VAD for each utterance ranging from 1 to 5. In our experiments, we linearly transform all VAD scores to the range $[0,1]$ during inference.

\subsection{Baseline Models}
We select the following baseline models for comparison:

\textbf{TL-ERC}~\cite{hazarika2021conversational}: The method pre-trains an encoder-decoder architecture on large-scale conversation data, then the weights of the encoder are transferred to ERC. 

\textbf{BERT-Large}~\cite{devlin-etal-2019-bert}: Following the standard pretraining-finetuning paradigm, the model is initialised from the pre-trained weights of BERT-Large, then a linear classifier follows the output of the BERT encoder for ERC task. 

\textbf{DialogXL}~\cite{Shen_Chen_Quan_Xie_2021}: Based on the XLNet~\cite{yang2019xlnet}, this work combines four types of dialogue-aware self-attention (global self-attention, local self-attention, speaker self-attention, listener self-attention) to model inter- and intra-speaker dependencies and proposes an utterance recurrence mechanism to model the long-range contexts.

\textbf{DAG-ERC}~\cite{shen-etal-2021-directed}: Utilising RoBERTa-Large~\cite{liu2019roberta} as the context-independent utterance encoder, this model builds a directed acyclic graph on the dialogue and uses a multi-layer graph neural network to aggregate the information on the graph. The node representations are utilised for ERC classification.

\textbf{SKAIG}~\cite{li-etal-2021-past-present}: This work extracts psychological commonsense knowledge and builds a graph on the dialogue according to different aspects of the knowledge. The corresponding knowledge representations are used to enhance the edge representations.

\textbf{COSMIC}~\cite{ghosal-etal-2020-cosmic}: This work uses the RNN to model the dialogue history for each participant and the context information. It also extracts utterance-level commonsense knowledge to model the speakers’ mental states and attentively infuses the knowledge into the utterance representations.

\textbf{Dis-VAE}~\cite{DBLP:conf/aaai/OngSCLNLSLW22}: This work utilises a VAE to model discourse information between utterances in an unsupervised manner and combines the learnt latent representations to the ERC encoder.

\textbf{SGED}~\cite{bao2022speaker}: This method proposes a speaker-guided encoder-decoder framework for ERC to model the intra- and inter-speaker dependencies in a dynamic manner. 

\textbf{CoG-BART}~\cite{li2021contrast}: Based on BART-Large, this work utilises contrastive learning and a response generation task to enhance utterance representations.

\textbf{SPCL}~\cite{song-etal-2022-supervised}: This work proposes a supervised prototypical
contrastive learning to improve performance on class-imbalanced data. Curriculum learning is also utilised to further enhance the model.

\textbf{CoMPM}~\cite{lee-lee-2022-compm}: This work proposes a context embedding module and a speaker-aware memory module to efficiently model the conversation and utterance-speaker relations.

\subsection{Implementation Details}
We conduct all experiments using a single Nvidia Tesla A100 GPU with 80GB of memory. We initialise the pre-trained weights of all PLMs and use the tokenisation tools both provided by Huggingface~\cite{wolf2019huggingface}. We leverage AdamW optimiser~\cite{DBLP:conf/iclr/LoshchilovH19} to train the model. We use the weighted-F1 measure as the evaluation metric for MELD and IEMOCAP. Since “neutral” occupies most of DailyDialog, we use micro-F1 for this dataset and ignore the label “neutral” when calculating the results as in the previous works~\cite{shen-etal-2021-directed,li2021contrast}.

The BART decoder brings about 1/3 more parameters to VAD-VAE than most of the selected baselines. Therefore, we also provide the results with a single-layer uni-directional LSTM utterance decoder to reduce the parameters and provide a more fair comparison. All hyper-parameters are tuned on the validation set. We also provide more implementation details: The batch size of experiments on all datasets is set to 4 except in DailyDialog, which is 16. Each iteration of training takes less than 0.75 GPU hours. We use a linear warm-up learning rate scheduling~\cite{goyal2017accurate} of warm-up ratio 20$\%$ and a peak learning rate 1e-5. We set a dropout rate of 0.1 and an L2-regularisation rate of 0.01 to avoid over-fitting. We do hyper-parameter search and determine $D_{(V)}=D_{(A)}=D_{(D)}=64$, $D_{(C)}=832$, $D_h$ = 1024, $\mu_E=0.8$, $\mu_I=1.0$. $\mu_{MI}$ is tuned between $[0.001, 0.01]$ on each validation set. More hyper-parameters will be listed within the source code. The names and versions of relevant software libraries and frameworks will be described in the source code. All reported results are averages of five random runs.

\begin{table}[h]
\caption{Test results on IEMOCAP, MELD and DailyDialog. ``-vCLUB'' denotes VAD-VAE trained without the vCLUB loss $\mathcal{L}_{MI}$. VAD-VAE$_{LSTM}$ denotes the results with an LSTM utterance decoder. VAD-VAE$_{H}$ denotes the results using the human-labeled utterance-level VAD scores. The numbers with $*$ indicate that the improvement over all baselines is statistically significant with p $<$ 0.05 under t-test. Best values are highlighted in bold.}
\label{tab:results}
\begin{center}
\begin{tabular}{l|cccc}
\toprule Model & IEMOCAP & MELD & DailyDialog\\ \midrule
TL-ERC & 59.30 & 57.46 & 52.46\\
BERT-Large & 60.98 & 61.50 & 54.09 \\
DialogXL & 65.94 & 62.41 & 54.93\\ \midrule
COSMIC & 65.28 & 65.21 & 58.48 \\
Dis-VAE & 68.23 & 65.34 & 60.95\\
SGED & 68.53 & 65.46 & --\\ 
CoMPM & 66.33 & 66.52 & 60.34\\ \midrule
DAG-ERC & 68.03 & 63.65 & 59.33\\
SKAIG & 66.96 & 65.18 & 59.75\\
CoG-BART & 66.18 & 64.81 & 56.29\\
SPCL & 69.74 & \textbf{67.25} & --\\
\midrule
VAD-VAE & \textbf{70.22}*($\pm$0.85) & 64.96($\pm$0.19) & \textbf{62.14}*($\pm$0.23)\\
\ \ -vCLUB & 69.19($\pm$0.66) & 65.94($\pm$0.31) & 61.23*($\pm$0.77)\\
VAD-VAE$_{LSTM}$ & 69.92 & 64.6 & 61.96*\\
\ \ -vCLUB & 69.72 & 64.89 & 61.4*\\
VAD-VAE$_{H}$ & 70.2* & -- & --\\

\bottomrule
\end{tabular}
\end{center}
\end{table}

\section{Results and Analysis}
\subsection{Overall Performance}
We present the performance of VAD-VAE and the baseline models on the three benchmark datasets in Table \ref{tab:results}. According to the results, BERT-Large and DialogXL outperform TL-ERC on all datasets, showing the advantages of PLM-based methods over RNN-based models that pre-train from scratch. Following this trend, the rest of the baseline models and our VAD-VAE all utilise RoBERTa as the utterance encoder (except CoG-BART which uses BART). COSMIC explicitly introduces mental state information to enrich the contexts, and the performance improves significantly on simple-context datasets MELD and DailyDialog. Dis-VAE, SGED, and CoMPM implicitly introduce speaker-related information and achieve over 68\% on IEMOCAP. To enhance the utterance representations, DAG-ERC and SKAIG build dialogue-level graphs to introduce priors on context modelling and perform well on all datasets. The competitive performance of CoG-BART and SPCL also proves the effectiveness of contrastive learning and response generation.

Overall, VAD-VAE achieves new state-of-the-art performance 70.22\% on IEMOCAP, 62.14\% on DailyDialog, and very competitive performance 65.94\% on MELD. Our model reaches an impressive 4.04\% improvement on IEMOCAP and a 5.85\% improvement on DailyDialog over CoG-BART, showing the advantage of VAE-based target utterance reconstruction over response generation. Test results also show that VAD-VAE outperforms Dis-VAE on all three datasets. We notice that Dis-VAE only builds a VAE to model the discourse information, while VAD-VAE uses VAE to reconstruct the target utterance directly. The results prove that target utterance reconstruction is appropriate to leverage the full potential of VAE. In addition, our method introduces NRC-VAD information and outperforms several strong knowledge-enhanced methods. For example, VAD-VAE outperforms COSMIC by over 4\% on both IEMOCAP and DailyDialog, which also utilises RoBERTa-Large as the utterance encoder and introduces mental state knowledge. These advantages reflect the effectiveness of the VAD supervision signals.

With the LSTM utterance decoder, VAD-VAE$_{LSTM}$ still outperforms previous state-of-the-art models on both IEMOCAP and DialyDialog. These results further prove the effectiveness of our proposed VAD-VAE structure under approximately the same number of parameters. In addition, VAD-VAE$_{LSTM}$ also achieves slightly worse performances on all datasets compared to VAD-VAE, which shows that a more powerful utterance decoder can benefit more on ERC. However, as VAD-VAE$_{LSTM}$ is more efficient in training, we recommend using the LSTM utterance decoder in most cases.

We can only test VAD-VAE$_{H}$ on IEMOCAP since all other datasets do not provide human-labeled utterance-level VAD scores. According to the results, VAD-VAE$_{H}$ only achieves a comparable performance of 70.2\% with VAD-VAE, which does not correspond with our hypothesis that fine-grained utterance-level VADs would provide more useful information. We notice that NRC-VAD follows strict best-worst scaling annotation and aggregation processes with a minimum of 6 annotators per word. In comparison, the IEMOCAP VAD annotation process follows a rough rule with two annotators for each utterance, which can bring inaccuracy to many labels and weaken the advantage of context-dependent VAD labels.

\begin{table}[h]
\caption{Results of ablation study. Best values are highlighted in bold.}
\label{tab:ablation}
\begin{center}
\resizebox{.47\textwidth}{!}{
\begin{tabular}{l|cccc}
\toprule Method & IEMOCAP & MELD & DailyDialog\\ \midrule
VAD-VAE & \textbf{70.22} & 64.96($\downarrow$0.98) & \textbf{62.14} \\
\ -vCLUB & 69.19($\downarrow$1.03) & \textbf{65.94} & 61.23($\downarrow$0.91)\\
\ -VAE Decoder& 67.92($\downarrow$2.30) & 63.99($\downarrow$1.95) & 60.16($\downarrow$1.98)\\
\midrule
\ -V,\ A,\ D Sup.& 66.85($\downarrow$3.37) & 63.99($\downarrow$1.95) & 61.01($\downarrow$1.13)\\
\ -V Sup.& 67.60($\downarrow$2.62) & 64.20($\downarrow$1.74) & 61.92($\downarrow$0.22)\\
\ -A Sup.& 68.06($\downarrow$2.16) & 64.59($\downarrow$1.35) & 61.38($\downarrow$0.76) \\
\ -D Sup.& 67.16($\downarrow$3.06) & 64.03($\downarrow$1.91) & 61.66($\downarrow$0.48)\\  \midrule
Utter. Encoder & 66.52($\downarrow$3.70) & 63.7($\downarrow$2.24) & 59.80($\downarrow$2.34)\\
\bottomrule
\end{tabular}}
\end{center}
\end{table}

\subsection{Ablation Study}
To investigate the effect of each module, we provide ablation analysis in Table \ref{tab:ablation}. ``-" denotes removing a module. ``vCLUB" denotes the MI minimisation modules. ``VAE Decoder" denotes the VAE decoder module for target utterance reconstruction. ``V Sup.", ``A Sup.", and ``D Sup." denote the NRC-VAD supervision signals corresponding to Valence, Arousal, and Dominance. ``Utter. Encoder" directly trains an ERC model on the context-aware utterance encoder.

According to the results, VAD-VAE achieves comparable performance with ``-vCLUB", which corresponds with our early hypothesis. With vCLUB loss added, the performance improves on IEMOCAP and DailyDialog. A possible reason is that MI minimisation enforces the model to learn dissimilar representations for VAD, which corresponds with the orthogonal nature of VAD in psychology. Without the VAE Decoder, the performance drops significantly on all datasets, which further indicates the effectiveness of the target utterance reconstruction task.

``-V, A, D Sup." leads to significant drops in all datasets, which proves our hypothesis that NRC-VAD supervision signals also provide fine-grained information to enhance ERC performance. In further comparisons of separately removing supervision signals for Valence, Arousal, and Dominance, the performance drops most with ``D Sup." removed for IEMOCAP and MELD and ``A Sup." removed for DailyDialog. We notice that the sentiment polarity of emotions is primarily determined by Valence, and similar emotions mainly differ in Arousal and Dominance. Therefore, these results show that our model benefits more from fine-grained information from Arousal and Dominance to distinguish similar emotions.




\begin{figure}[htpb]
\centering
\includegraphics[width=8cm,height=4.8cm]{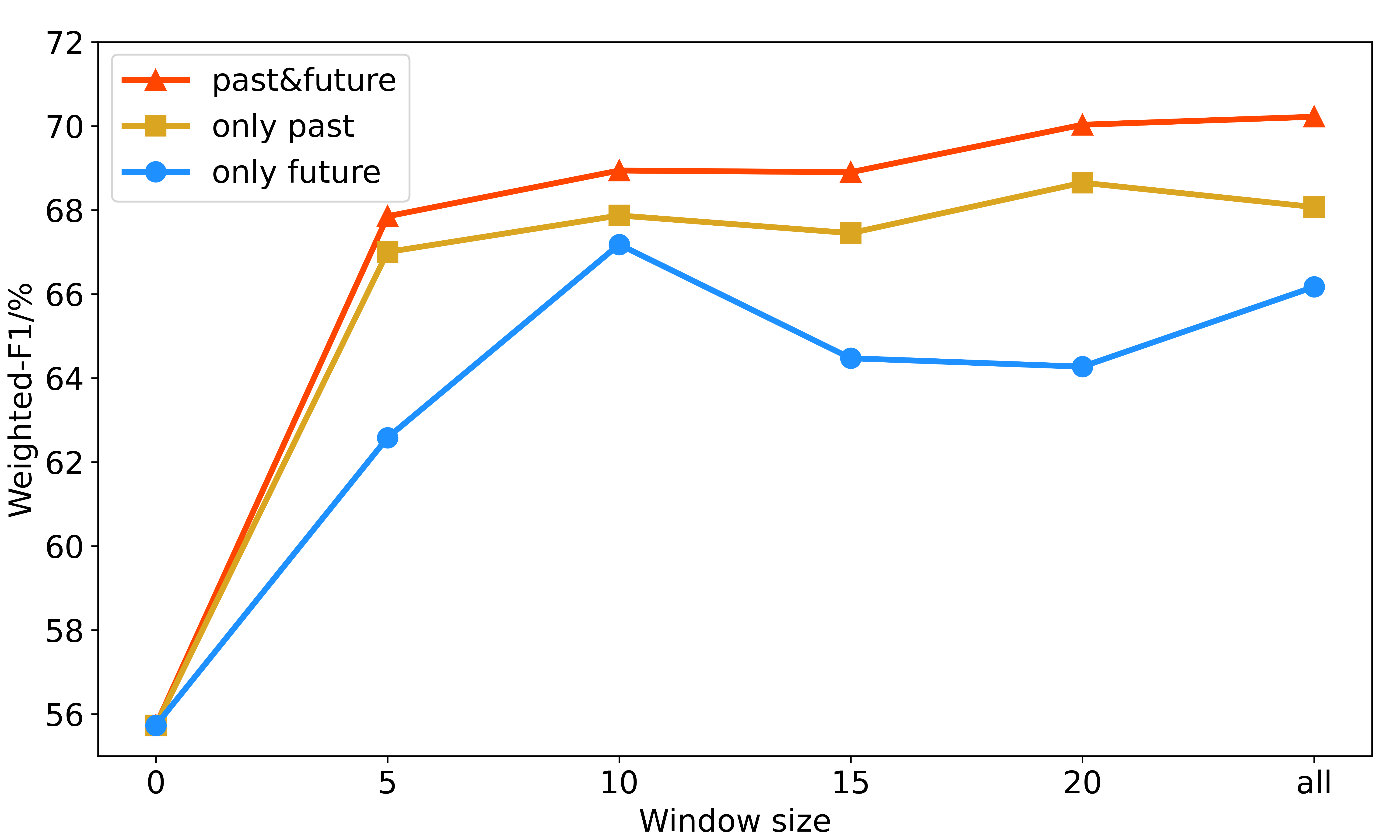}
\caption{The performance of VAD-VAE on IEMOCAP with different context window sizes in the input sequence. We separately present results with both past and future contexts and only past/future contexts. "all" denotes setting the corresponding window sizes as wide as possible.}
\label{fig:context}
\end{figure}

\subsection{Context Influence}
In previous experiments, we set both past and future context window sizes $W_p$ and $W_f$ as wide as possible (the maximum input length of RoBERTa encoder is 512) to provide more contextual information. In Figure \ref{fig:context}, we further present VAD-VAE performances with different context window sizes to investigate the influence of multi-range contextual information and guide future research directions on context modeling.

According to the results, both past and future contexts significantly enhance the model performance compared to no context (window size 0), but $past\&future$ further outperforms both \emph{only past} and \emph{only future} on all window sizes, which shows that VAD-VAE benefits from combined emotional reasoning with past and future contexts. In addition, though the nearest 5 contexts provide most of the improvement (over 10\%) for $past\&future$ and \emph{only past}, their performances continuously increase with wider window sizes, proving that long-range conversational history also provides useful information. On the other hand, the performance of \emph{only future} reaches the top with a window size of 10 but decreases with longer contexts. These results denote near-future contexts as more indicative of emotional cues for the target utterance, while unrelated longer future contexts can bring noises to the emotional reasoning.

\begin{table*}[h]
\caption{The Pearson’s correlation coefficients between predicted VAD scores and supervision signals on all test sets, and the average vCLUB estimates of MI between VAD latent distributions. Best values are highlighted in bold.}
\label{tab:pcc}
\begin{center}
\begin{tabular}{l||cccc|cccc|cccc}
\toprule
 & \multicolumn{4}{c|}{IEMOCAP} & \multicolumn{4}{c|}{MELD} & \multicolumn{4}{c}{DailyDialog}\\
\midrule
Methods & V & A & D & MI & V & A & D & MI & V & A & D & MI\\
\midrule
$\mathcal{L}_{ELBO}$ & 0.434 & 0.156 & 0.094 & 0.732 & 0.021 & 0.133 & 0.152 & 0.398 & 0.071 & 0.029 & -0.017 & 0.356\\
\ +$\mathcal{L}_{MI}$ & 0.306 & 0.132 & 0.150 & \textbf{0.133} &0.044 & 0.230 & 0.141& \textbf{0.068} & 0.115 & -0.026 & -0.016 & \textbf{0.094}\\
\ +$\mathcal{L}_{INFO}$ & \textbf{0.882} & 0.708 & 0.743& 0.531 & 0.565 & \textbf{0.643} & 0.557& 0.885 & 0.424 & 0.348 & 0.321 & 0.789\\
\ +$\mathcal{L}_{INFO}$+$\mathcal{L}_{MI}$ & 0.872 & \textbf{0.715} & \textbf{0.765}& 0.355 & \textbf{0.568} & 0.639 & \textbf{0.558}& 0.312 & \textbf{0.433} & \textbf{0.355} & \textbf{0.330} & 0.226\\
\bottomrule
\end{tabular}
\end{center}
\end{table*}

\begin{figure*}[htpb]
\centering
\includegraphics[width=18cm,height=4.5cm]{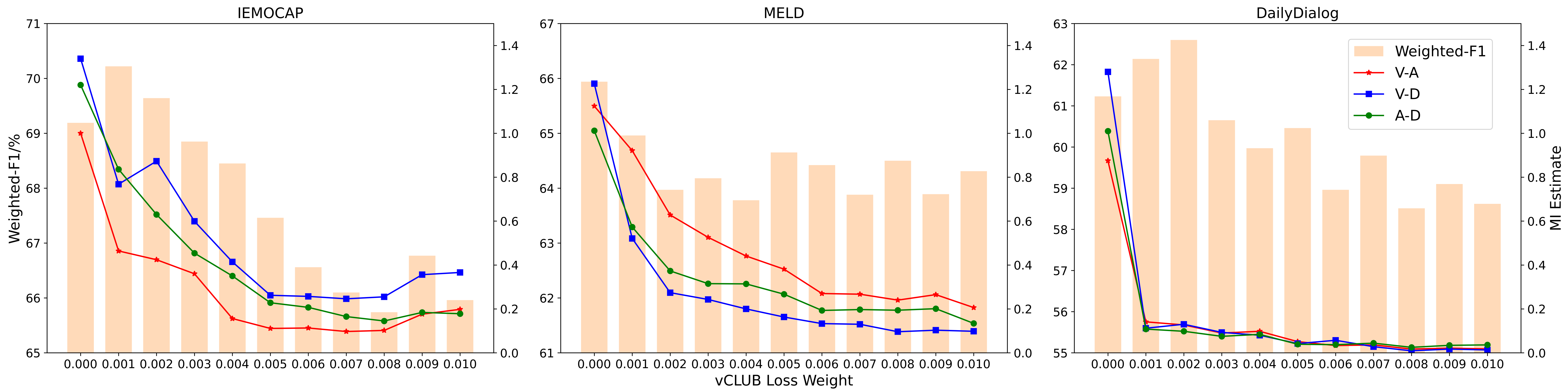}
\caption{The performance of VAD-VAE on ERC and the MI estimates between VAD pairs with different vCLUB weight coefficients $\mu_{MI}$ on the three test sets.}
\label{fig:tradeoff}
\end{figure*}

\subsection{Disentanglement Evaluation}
We analyse the effects of VAD supervision signals ($\mathcal{L}_{INFO}$) and MI minimisation ($\mathcal{L}_{MI}$) on enhancing VAD disentanglement. In Table \ref{tab:pcc}, we present the Pearson's correlation coefficients between the predicted VAD scores from latent representations and the supervision signals from NRC-VAD on all three test sets. Higher values indicate more precise predictions, denoting better \emph{Informativeness}. We also provide the average vCLUB estimates of MI between VAD latent distributions on each test set, with lower values denoting lower estimates of MI upper bounds and better \emph{Independence}.


\subsubsection{Informativeness}
According to the results, the model performs poorly on all datasets (Pearson's correlation coefficients below 0.2 in most cases) with VAE reconstruction loss ($\mathcal{L}_{ELBO}$) or $\mathcal{L}_{MI}$ introduced, since VAD features can be embedded in Content space without specific supervisions. We observe significant improvement in \emph{Informativeness} for VAD with $\mathcal{L}_{INFO}$, which brings over 0.5 Pearson's correlation coefficients gain for IEMOCAP and 0.3 for MELD and DailyDialog. These results reflect the effectiveness of NRC-VAD supervision signals. On top of $\mathcal{L}_{INFO}$, $\mathcal{L}_{MI}$ further improves the Pearson's correlation coefficients scores in most cases, which shows that MI minimisation also helps to enhance \emph{Informativeness} of VAD representations to some extent.

\subsubsection{Independence}
For all datasets, The vCLUB estimates remain high with only $\mathcal{L}_{ELBO}$ introduced since the unified distributed representation in VAE encourages strong correlations between each part. With $\mathcal{L}_{INFO}$, we observe even higher vCLUB in MELD and DailyDialog. In this case, our model is only optimised for \emph{Informativeness}, and enforces full utilisation of all latent spaces, which leads to high MI. Introducing only $\mathcal{L}_{MI}$ has the lowest vCLUB in all datasets. However, it achieves bad performance in \emph{Informativeness}. With both $\mathcal{L}_{MI}$ and $\mathcal{L}_{INFO}$, VAD-VAE not only achieves the best results in VAD predictions but also greatly decreases vCLUB compared with only $\mathcal{L}_{INFO}$, showing the satisfactory trade-off between \emph{Informativeness} and \emph{Independence}.

\subsubsection{ERC-Independence Trade-off}\label{appendix:tradeoff}
To further investigate the influence of MI minimisation on ERC performance, we present the results of tuning vCLUB loss coefficients $\mu_{MI}$ on the three test sets in Figure \ref{fig:tradeoff}. 

For all datasets, the MI estimates decrease as $\mu_{MI}$ increases, showing the effectiveness of the vCLUB method. Meanwhile, ERC performance decreases at acceptable rates. For IEMOCAP, with $\mu_{MI}$=0.005, all MI upper bounds steadily drop below 0.3 while keeping a competitive ERC performance: over 67\%. As $\mu_{MI}$ further increases, there is no apparent decrease in MI estimates. We have similar observations for MELD at $\mu_{MI}$=0.006. On the other hand, for DailyDialog, all MI estimates drop rapidly below 0.1 with small $\mu_{MI}$=0.001. According to the statistics in Table \ref{tab:statistics}, DailyDialog has more abundant training data, which provides rich information for understanding Valence, Arousal, and Dominance. Simpler contexts of DailyDialog can also facilitate VAD disentanglement. From the perspective of each VAD pair, DailyDialog achieves high disentanglement for all VAD pairs, while on the other two datasets, the performances are uneven. Valence-Dominance in IEMOCAP and Valence-Arousal in MELD remain above 0.2. These results show that the pair-wise difficulties of VAD disentanglement are possibly related to the complexity of the contexts.

\begin{figure}[htpb]
\centering
\includegraphics[width=7.8cm,height=7.8cm]{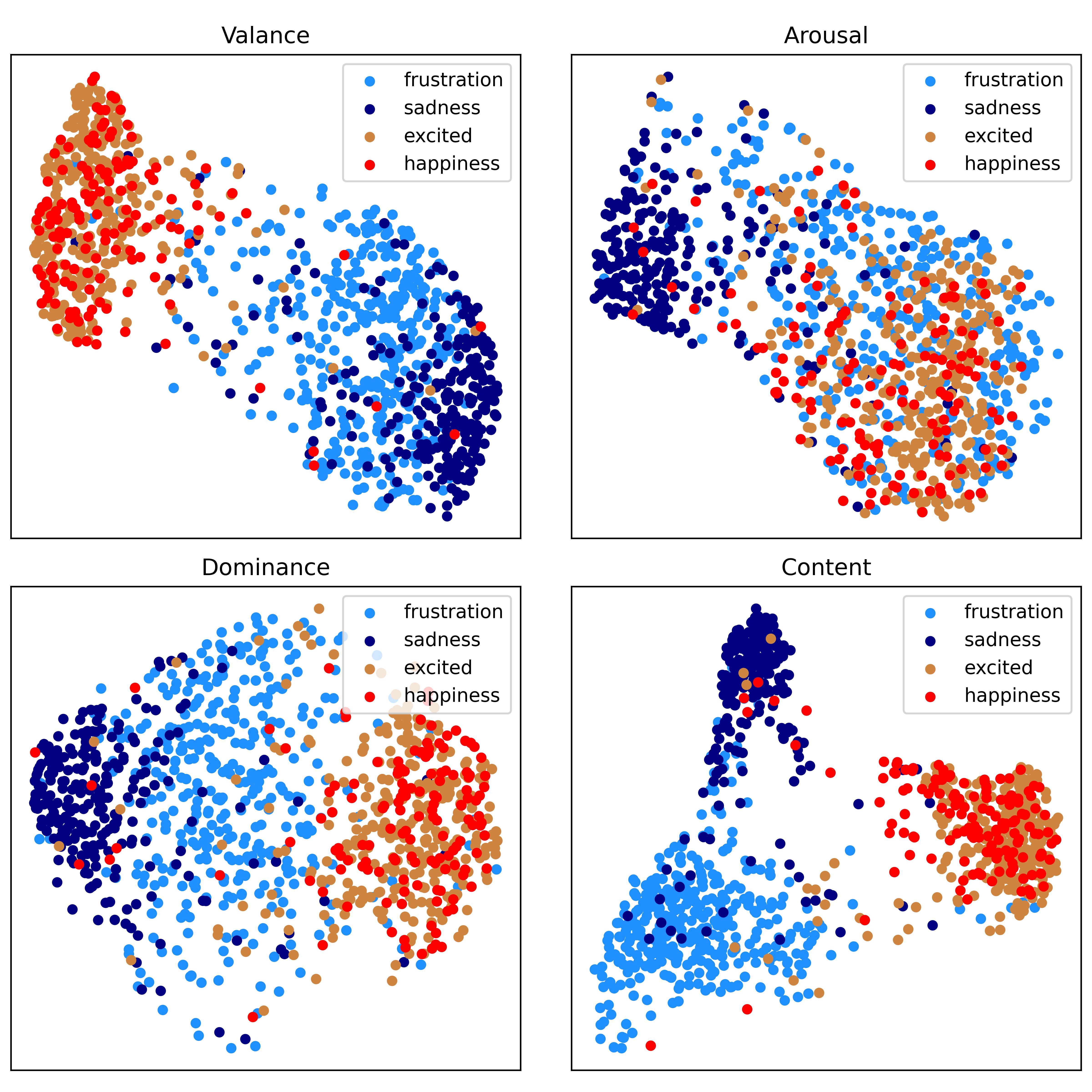}
\caption{UMAP visualisations of Valence, Arousal, Dominance and Content representations for IEMOCAP test results.}
\label{fig:scatter}
\end{figure}

\begin{table*}[h]
\caption{Four cases where target utterance reconstruction leverages the critical context of the target utterance for emotion reasoning, while the next utterance provides limited information. ``CoG-BART'', ``VAD-VAE'' and ``Human'' denotes the corresponding predictions for response generation (CoG-BART), target utterance reconstruction (VAD-VAE) and the human-annotated emotion labels. Key information in the context: bold.}
\label{tab:case}
\begin{center}
\resizebox{1.\textwidth}{!}{
\begin{tabular}{cccccccccccc|ccc}
\toprule
\multicolumn{4}{c}{Target Utterance} & \multicolumn{4}{c}{Next Utterance} & \multicolumn{4}{c|}{Context} & CoG-BART & VAD-VAE & Human\\ 
\midrule
\multicolumn{4}{c}{\begin{tabular}[c]{@{}l@{}}\#1 Chandler: It's all finished!\end{tabular}} & \multicolumn{4}{c}{\begin{tabular}[c]{@{}l@{}}Joey: This was Carol's favorite beer. She\\ always drank it out of can, I should have\\ known.\end{tabular}} & \multicolumn{4}{c|}{\begin{tabular}[c]{@{}l@{}}Chandler: That's OK, anyway.\\Joey: Tell Monica we are \textbf{done with}\\\textbf{the bookcase} at last!\end{tabular}} & \emph{neutral} & \emph{joyful} & \emph{joyful}\\
\midrule
\multicolumn{4}{c}{\begin{tabular}[c]{@{}l@{}}\#2 James: OK, that's OK. Improvise\\is just good.\end{tabular}} & \multicolumn{4}{c}{\begin{tabular}[c]{@{}l@{}}James: Well, what about we go to a great\\jazz show now, and um-\end{tabular}} & \multicolumn{4}{c|}{\begin{tabular}[c]{@{}l@{}}James: Well, you've got to tell me\\the details. What did he say?\\Joey: I don't-I don't know \textbf{how much}\\\textbf{plan} he had done.\end{tabular}} & \emph{neutral} & \emph{excited} & \emph{excited}\\
\midrule
\multicolumn{4}{c}{\begin{tabular}[c]{@{}l@{}}\#3 Mary: It's a good idea to live together\\before you get married.\end{tabular}} & \multicolumn{4}{c}{\begin{tabular}[c]{@{}l@{}}James: I think so, too. Are you going\\ to get a house or-\end{tabular}} & \multicolumn{4}{c|}{\begin{tabular}[c]{@{}l@{}}Linda: I have to think about moving now.\\We are going to \textbf{move in together}!\\ Monica: I thought you two already\\did that...\end{tabular}} & \emph{surprised} & \emph{happy} & \emph{happy}\\
\midrule
\multicolumn{4}{c}{\begin{tabular}[c]{@{}l@{}}\#4 Chandler: Hey thanks. Joey is the uh-\\fellow processor.\end{tabular}} & \multicolumn{4}{c}{\begin{tabular}[c]{@{}l@{}}Scott: No kidding.\end{tabular}} & \multicolumn{4}{c|}{\begin{tabular}[c]{@{}l@{}} Joey: But don't you need more experiences\\for a job like that?\\Chandler: It's not that hard to learn. Hey,\\\textbf{you're an actor, act like a processor.}\\James: Hey Chandler, this morning's\\projections here.\end{tabular}} & \emph{frustrated} & \emph{angry} & \emph{angry}\\
\bottomrule
\end{tabular}}
\end{center}
\end{table*}

\subsection{Disentangled Representations Visualisation}
To conduct a more interpretable analysis of the disentangled representations, we present the UMAP~\cite{mcinnes2018umap} visualisations of VAD representations in the IEMOCAP test set for four representative emotions in Figure \ref{fig:scatter}. Their corresponding NRC-VAD assignments are presented in Table \ref{vads}. As shown, for Valence and Dominance, positive and negative emotions are well separated, while emotions within one polarity overlap. In the visualisation of Arousal, ``happiness'', ``excited'' and ``frustration'' lie close while ``sadness” separates away. These observations correspond with the NRC-VAD assignments, indicating the quality of the learnt VAD representations. In addition, the distribution of each emotion shows continuity and completeness conditions.

\begin{figure}[htpb]
\centering
\includegraphics[width=8cm,height=3.14cm]{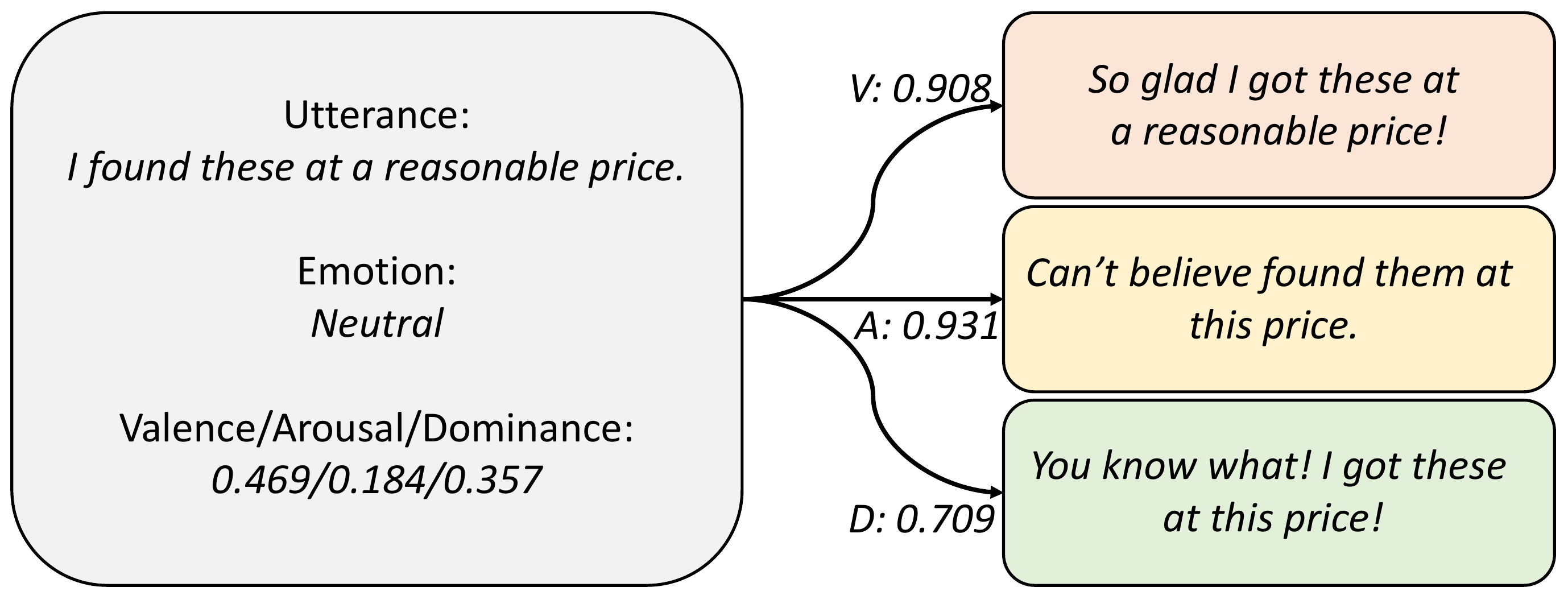}
\caption{A case where the target utterance is reconstructed via VAD-VAE by separately adjusting Valence, Arousal, and Dominance.}
\label{fig:generate case}
\end{figure}

We observe very regularised latent distributions for Content. The distribution of each emotion is close to a Gaussian distribution and shows continuity and completeness conditions for ``frustration'', ``sadness'' and ``happiness'' utterances. However, the representations of ``excited'' and ``happiness'' still overlap, possibly due to the similar way of expressing excited and happy emotions. Overall, with the well learnt disentangled VAD representations and sound properties of the content space, one direction of future work is to explore the potential of VAD-VAE in the controlled affective text generation task. Unlike previous works, which control categorical emotions, our work aims to control more fine-grained sentiments by adjusting Valence, Arousal, and dominance separately. To provide a more intuitive view, we provide a case of VAD-based controlled generation in Figure \ref{fig:generate case}. The VAD representations of the neutral target utterance are replaced by the VAD representations of a randomly sampled excited utterance. As shown, the utterance decoder can adjust the tones of the utterance according to the input VAD representations. For example, with the new 0.908 Valence representation, the reconstructed utterance "so glad I got these at a reasonable price!" shows more pleasantness. 

\begin{figure}[htpb]
\centering
\includegraphics[width=7.5cm,height=4.5cm]{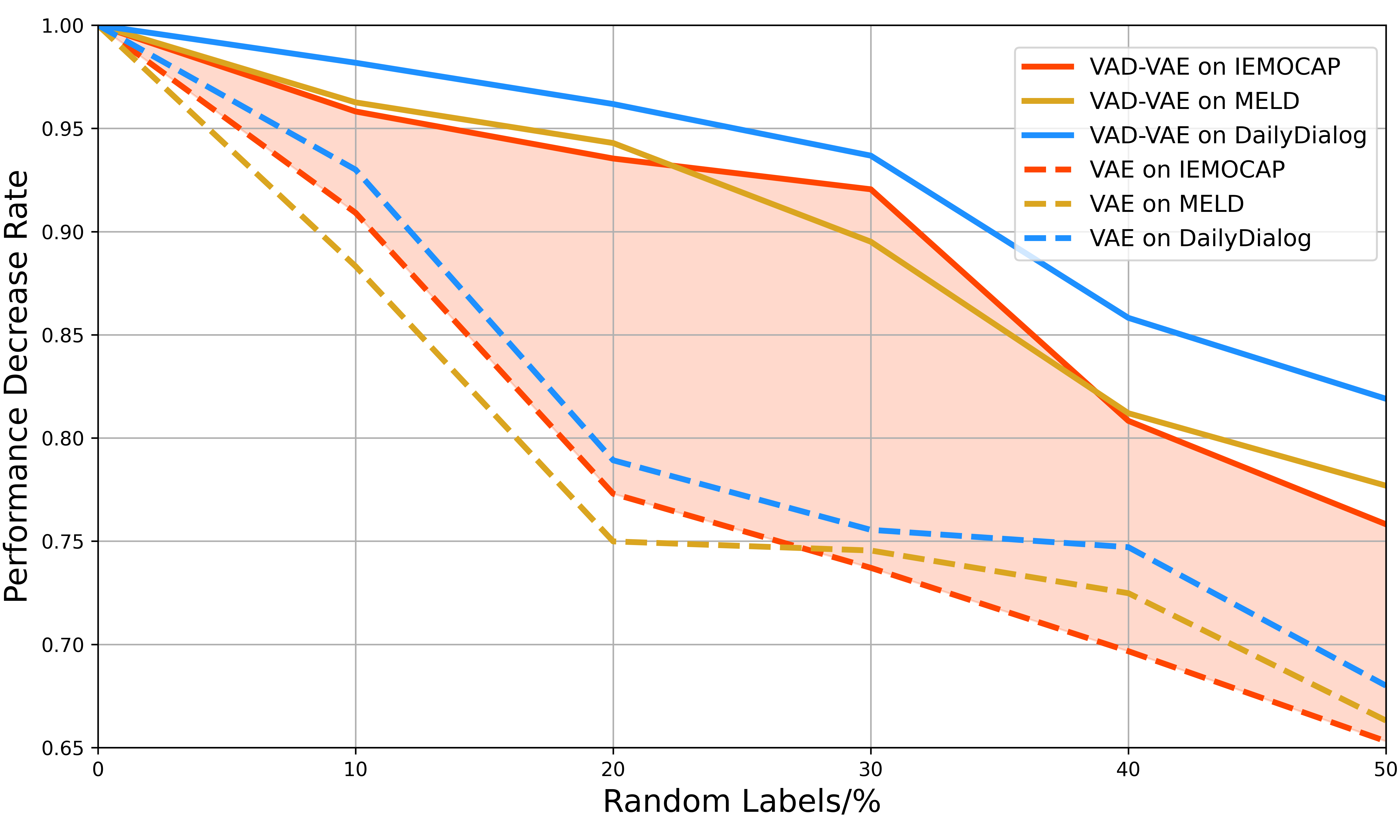}
\caption{Performance decrease rate of VAD-VAE and vanilla VAE on three ERC test sets with different random replacement percentages of training labels.}
\label{fig:robustness}
\end{figure}

\subsection{Robustness Evaluation}
In Figure \ref{fig:robustness}, we evaluate the robustness of disentangled representations by randomly replacing a percentage (0\% to 50\%) of training labels with false labels, then comparing the performance decrease rate (For $\alpha\%$ replacement, the rate is computed as $\frac{F1\ for\ \alpha\%\ replacement}{F1\ for\ 0\%\ replacement}$) of VAD-VAE and the vanilla VAE method. Higher rates indicate better performance and more robustness.

According to the results, disentangled VAD representations achieve higher performance decrease rates than entangled representations on all datasets. For example, VAD-VAE outperforms VAE by an average of 12.24\% with all label replacement percentages (orange shaded area). With 50\% of random label replacement, VAD-VAE retains over 75\% of performances on all test sets while the performances of VAE all fall below 70\%. These results show that VAD-VAE can tolerate a higher level of misinformation during training, indicating the robustness of disentangled VAD representations over the entangled representations. A possible reason is that disentangled representations are explicitly trained to extract VAD information, which provides helpful guidance during inference when other features are misleading.

\subsection{Case Study for Target Utterance Reconstruction}
Target utterance reconstruction enables VAD-VAE to learn long-range dependencies and outperforms the response generation-based model CoG-BART. We provide four cases from the test results of VAD-VAE and CoG-BART in Table \ref{tab:case} for further investigation.

According to our observation, there are mainly two scenarios where response generation provides limited information and mispredicts the emotion: (a) Change of the discussion topic in the response. In the next utterance of case \#1, the discussion topic changes from ``Finish assembling the bookcase'' to ``Carol's favorite beer''. In the next utterance of case \#2, the topic changes from ``His plan'' to ``Go to a great jazz show''. In these cases, emotion-related clues in the response are irrelevant to the target utterance. (b) Indirect response to the current utterance. In case \#3, the next utterance merely expresses agreement with the target utterance and expands the dialogue. In case \#4, the response is simply sarcasm. These responses provide crucial clues for future utterances but little information for the target utterance. These scenarios are more common in multi-party dialogues. In contrast, VAD-VAE can learn complex inter- and intra-speaker influences since it learns critical information via the context-aware target utterance reconstruction. For example, VAD-VAE focuses on the previous discussion topic ``done with the bookcase''(case \#1) and ``his plan''(case \#2). It can also extract key information from long-range contexts, such as in cases \#3 and \#4. These advantages allow VAD-VAE to make correct emotion predictions in these cases.

\section{Conclusion}
This paper proposes a VAD-disentangled Variational Autoencoder for emotion recognition in conversations. We first introduce an auxiliary target utterance reconstruction task via the VAE framework. Then we disentangle three key features: Valence, Arousal, and Dominance, from the latent space. VAD supervision signals and a mutual information minimisation task are also utilised to enhance the disentangled VAD representations. 

Experiments show that VAD-VAE outperforms the state-of-the-art model on two ERC datasets. Ablation and case studies prove the effectiveness of the proposed VAE-based target utterance reconstruction task and VAD supervision signals. Further quantitative analysis and visualisation also show that VAD-VAE learns disentangled VAD representations with informativeness, independence and robustness. In the future, we will leverage these decent VAD representations to explore fine-grained emotion control for affective text generation by adjusting Valence, Arousal, and Dominance separately.


%

\ifCLASSOPTIONcompsoc
  \section*{Acknowledgments}
\else
  \section*{Acknowledgment}
\fi

This work was supported in part by the Alan Turing Institute, UK. This work was also supported by the University of Manchester President’s Doctoral Scholar award.

\ifCLASSOPTIONcaptionsoff
  \newpage
\fi



\bibliographystyle{IEEEtran}
\bibliography{egbib}
%

%

\vskip 0pt plus -1fil

\begin{IEEEbiography}[{\includegraphics[width=1in,height=1.25in,clip,keepaspectratio]{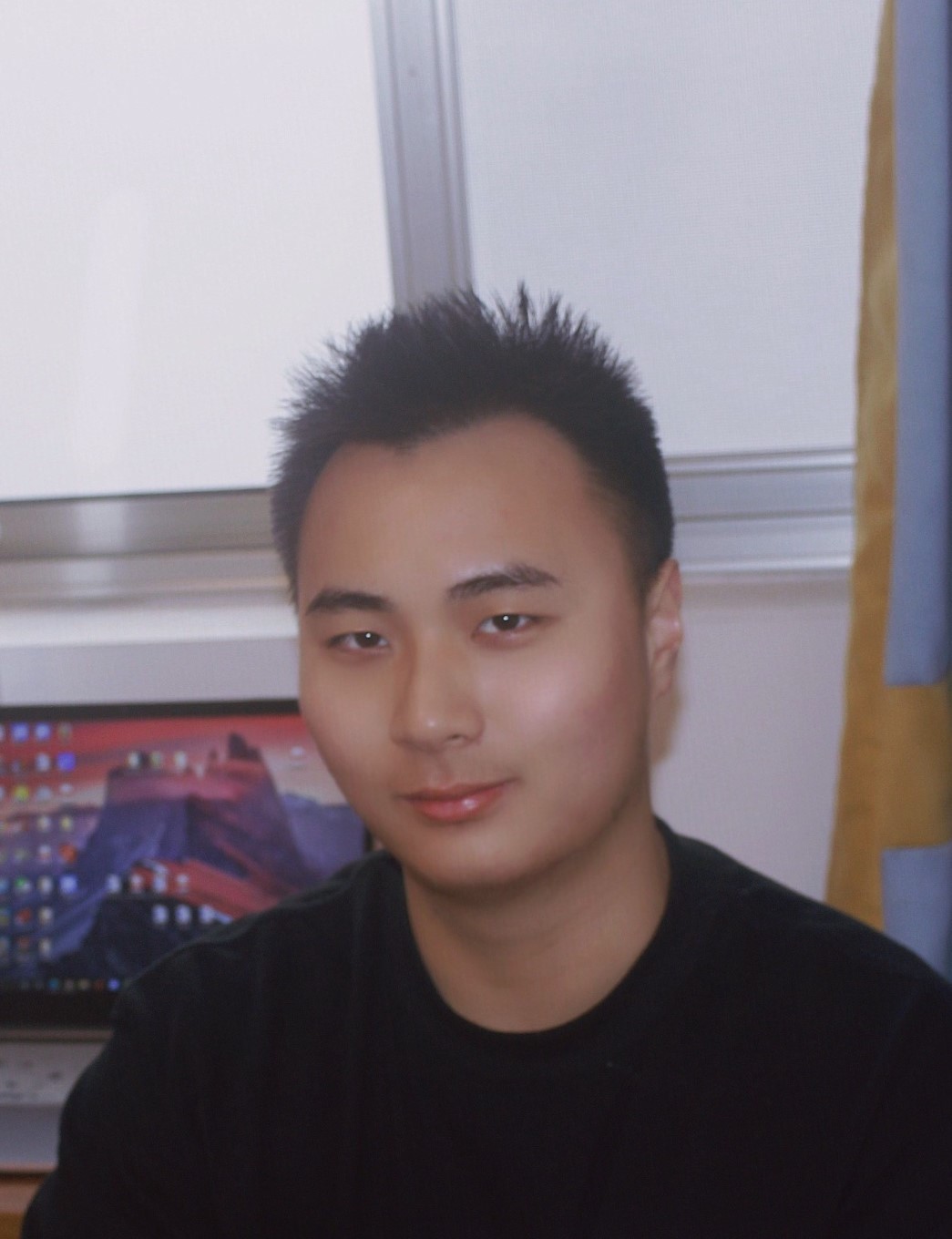}}]{Kailai Yang} received the B.E. degree in computer science and technology from Harbin Institute of Technology, China, in 2021. He is currently pursuing the Ph.D degree in computer science at the University of Manchester, United Kingdom. His current research interests include natural language processing, affective computing and deep learning.
\end{IEEEbiography}

\vskip 0pt plus -1fil

\begin{IEEEbiography}[{\includegraphics[width=1in,height=1.25in,clip,keepaspectratio]{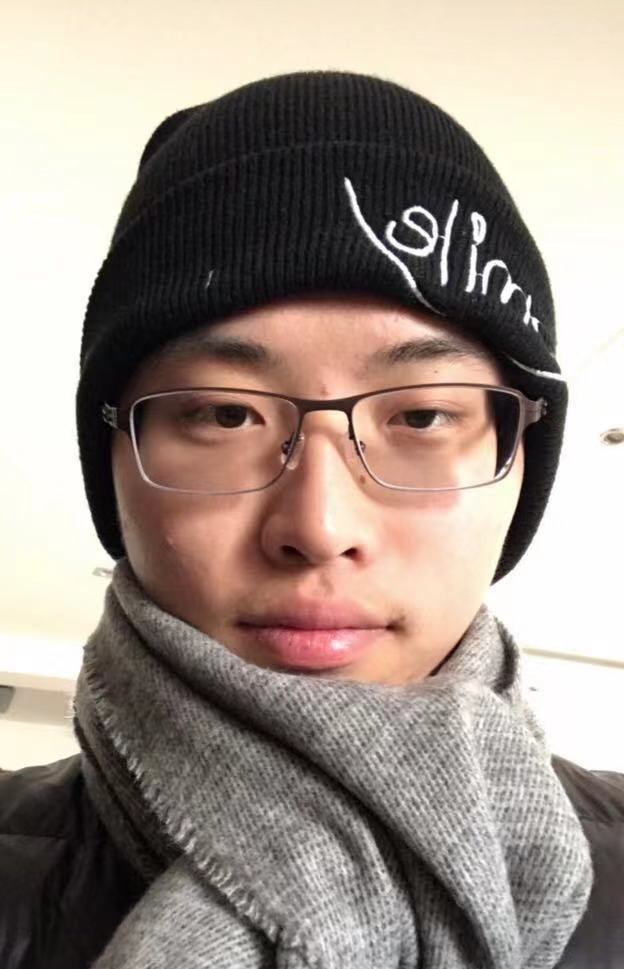}}]{Tianlin Zhang} received the B.S. degree in software
engineering from Nankai University, China, in 2017 and the M.S. degree in computer science from University of Chinese Academy of Sciences, China, in 2020. He is currently pursuing the Ph.D. degree in computer science at the University of Manchester, United Kingdom. His current research interests include natural language processing, affective computing and deep learning.
\end{IEEEbiography}

\vskip 0pt plus -1fil

\begin{IEEEbiography}[{\includegraphics[width=1in,height=1.25in,clip,keepaspectratio]{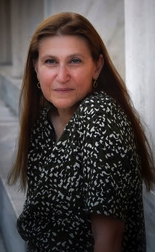}}]{Sophia Ananiadou} is the Director of the UK
National Centre for Text Mining (NaCTeM) and a
Professor of Computer Science with The University of Manchester. She is also a Turing Fellow in the Alan Turing Institute and a visiting professor
in AIST/AIRC. Her main research contributions
are in the areas of biomedical text mining and
natural language processing.
\end{IEEEbiography}




\end{document}